\documentclass[letterpaper, 10 pt, conference]{ieeeconf}
\IEEEoverridecommandlockouts
\usepackage[bookmarks=true]{hyperref}

\usepackage{url}

\usepackage{amsmath,amsfonts}
\usepackage{array}
\usepackage[caption=false,font=normalsize,labelfont=sf,textfont=sf]{subfig}
\usepackage{textcomp}
\usepackage{stfloats}
\usepackage{tabularray}
\usepackage{tabu}
\usepackage{booktabs} % for professional tables
\usepackage{multirow}
\usepackage{multicol}
\usepackage{adjustbox}

\usepackage{xcolor}
\usepackage{xspace}
\usepackage{ifthen}

\usepackage{graphicx}
\usepackage{tikz}
\usetikzlibrary{positioning}

\let\labelindent\relax %needed for IEEE conflict with enumitem
\usepackage{enumitem} % for \begin{description}[align=left]

\usepackage{amsmath,amssymb,amsfonts}
\usepackage{dsfont}
\usepackage{color}
\usepackage{balance}
\usepackage{gensymb}
\usepackage{siunitx} 
\usepackage{algorithm}
\usepackage{algorithmicx}
\usepackage{algpseudocode}
\usepackage{alphalph}
\usepackage{comment}
\usepackage{soul}
\usepackage{alphalph,etoolbox}

\graphicspath{{figures/}}

\allowdisplaybreaks[4]

\definecolor{navy}{rgb}{0,0,0.5}
\definecolor{dgreen}{rgb}{0,.7,0.6}
\definecolor{dyellow}{rgb}{.7,.7,0}
\definecolor{dred}{rgb}{1,0,0}
\definecolor{dblue}{rgb}{0,0,0.7}
\definecolor{brightblue}{rgb}{0.,0.5,1} 
\definecolor{dorange}{rgb}{0.9,0.5,0.1}
\definecolor{dgray}{rgb}{0.5,0.5,0.5}

\usepackage{diagbox}

\patchcmd{\subequations}{\alph{equation}}{\alphalph{\value{equation}}}{}{}

\algrenewcommand\algorithmicrequire{\textbf{Input:}}
\algrenewcommand\algorithmicensure{\textbf{Output:}}

\usepackage[utf8]{inputenc} % allow utf-8 input
\usepackage[T1]{fontenc}    % use 8-bit T1 fonts
\usepackage{url}            % simple URL typesetting
\usepackage{booktabs}       % professional-quality tables
\usepackage{amsfonts}       % blackboard math symbols
\usepackage{nicefrac}       % compact symbols for 1/2, etc.
\usepackage{microtype}      % microtypography
\usepackage{tabularx}
\usepackage{microtype}
\usepackage{graphicx}
\usepackage{subcaption}

\usepackage[utf8]{inputenc} % allow utf-8 input
\usepackage[T1]{fontenc}    % use 8-bit T1 fonts
\usepackage{dsfont}
\usepackage{nicefrac}       % compact symbols for 1/2, etc.
\usepackage{color}
\usepackage{mathtools}
\usepackage{amsmath,amssymb}
\usepackage{bm}
\usepackage{siunitx}
\usepackage{wrapfig}
\usepackage{lipsum}
\usepackage{makecell}
\usepackage{subcaption}
\usepackage[capitalise, nameinlink]{cleveref}

\usepackage{footnote}
\makesavenoteenv{tabular}
\makesavenoteenv{table}

\usepackage{amsmath}
\usepackage{amssymb}
\usepackage{mathtools}
\usepackage{pifont}% http://ctan.org/pkg/pifont
\newcommand{\cmark}{\ding{51}}%
\usepackage{graphicx} % for pdf, bitmapped graphics files
\usepackage{subcaption}

\usepackage{hyperref}
\usepackage{booktabs}
\usepackage{xspace}
\makeatletter
    \let\NAT@parse\undefined
\makeatother

\usepackage{hyperref}
\hypersetup{
    colorlinks=true,
    linkcolor=blue,
    filecolor=magenta,      
    urlcolor=magenta,
    citecolor=orange,
}
\usepackage[capitalise, nameinlink]{cleveref}

\usepackage[
  backend=biber,
  style=ieee,          % IEEE-like numeric style
  natbib=true,         % keep \citet/\citep working like natbib
  maxcitenames=2,      % truncate citations after 2 authors -> et al.
  mincitenames=1,
  maxbibnames=10,       % in the reference list, show up to 10 authors
  minbibnames=1,       % then truncate to "FirstAuthor et al."
  sorting=none         % order of appearance (IEEE)
]{biblatex}

\usepackage{eso-pic}

\newcommand\IEEEcopyrightnotice{%
  \begin{minipage}{0.95\textwidth}
  \footnotesize
  \copyright~2026 IEEE. Personal use of this material is permitted. Permission from IEEE must be obtained for all other uses, in any current or future media, including reprinting/republishing this material for advertising or promotional purposes, creating new collective works, for resale or redistribution to servers or lists, or reuse of any copyrighted component of this work in other works.
  \end{minipage}
}
\newcommand{\algoname}{MonoDuo\xspace}
\newcommand{\paragraphc}[1]
{\vspace{0.1em}\noindent\textbf{#1}}

\title{\LARGE \bf
MonoDuo: Using One Robot Arm to Learn Bimanual Policies
}

\author{$^{\dagger}$Sandeep Bajamahal$^{1*}$, Lawrence Yunliang Chen$^{1*}$, Toru Lin$^{1}$, Zehan Ma$^{1}$, Jitendra Malik$^{1}$, Ken Goldberg$^{1}$% <-this % stops a space
\thanks{$^{1}$University of California, Berkeley
        }%
\thanks{$^{\dagger}$Corresponding author: sandeep24@berkeley.edu}}

\begin{document}

\makeatletter
\let\@oldmaketitle\@maketitle% Store \@maketitle
\renewcommand{\@maketitle}{\@oldmaketitle% Update \@maketitle to insert...
\vspace{-.05cm}
  \begin{center}
  \includegraphics[width=\textwidth]{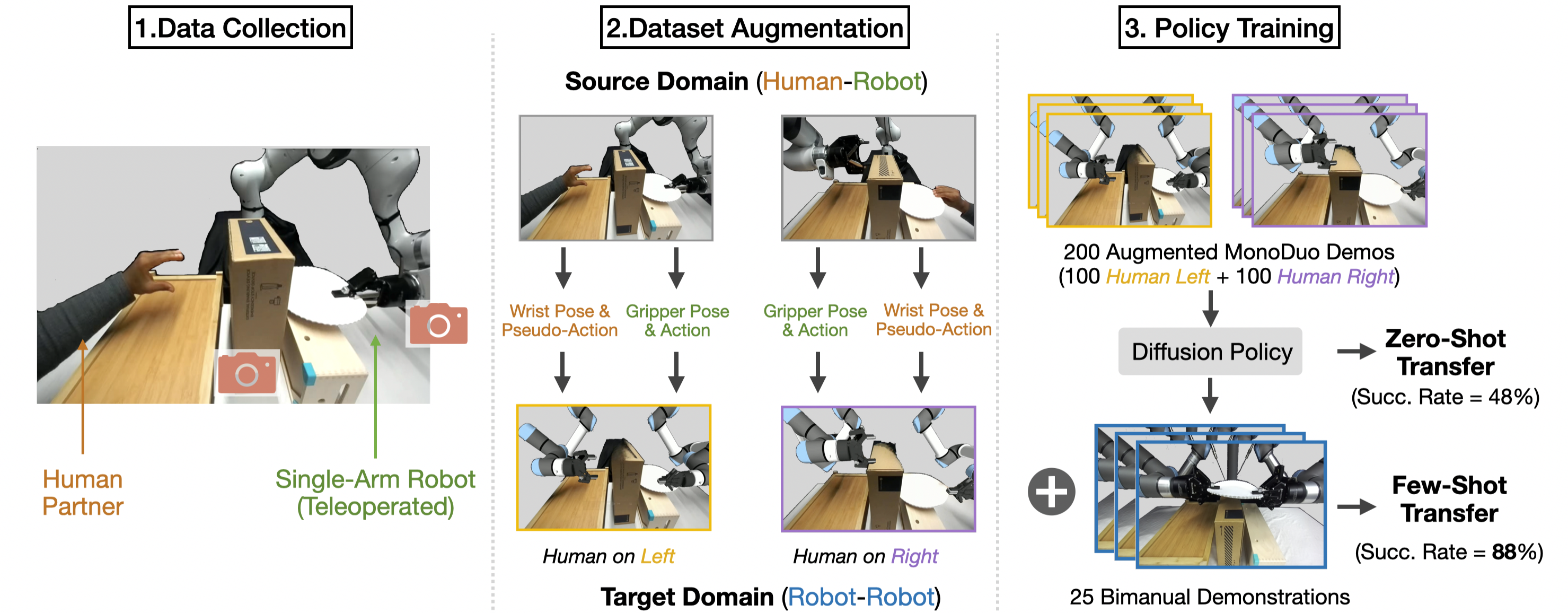}
    \captionof{figure}{\textbf{Overview of MonoDuo.} The teleoperation system uses a fixed RGB-D camera and a wrist-mounted camera. We begin by teleoperating a single-arm robot to collaborate with a human arm on a bimanual task, alternating left-right arm roles across episodes. This results in complementary interaction data covering both sides of the task. These human-robot bimanual demonstrations are then augmented into synthetic robot-robot bimanual demonstrations using segmentation and inpainting techniques, creating a visually and physically grounded dataset for training bimanual robots.}
    \label{fig:overview}
    \vspace{-1em}
    \end{center}
}
% \addtocounter{figure}{-1}
\makeatother

\maketitle

\AddToShipoutPictureBG*{%
  \AtPageLowerLeft{%
    \hspace{0.35in}%
    \raisebox{0.25in}{\IEEEcopyrightnotice}%
  }%
}

\thispagestyle{empty}
\pagestyle{empty}

%%%%%%%%%%%%%%%%%%%%%%%%%%%%%%%%%%%%%%%%%%%%%%%%%%%%%%%%%%%%%%%%%%%%%%%%%%%%%%%%
\begin{abstract}
Bimanual coordination is essential for many real-world manipulation tasks, yet learning bimanual robot policies is limited by the scarcity of bimanual robots and datasets. Single-arm robots, however, are widely available in research labs. Can we leverage them
to train bimanual robot policies? We present \textit{MonoDuo}, a framework for learning bimanual manipulation policies using single-arm robot demonstrations paired with human collaboration. MonoDuo collects data by teleoperating a single-arm robot to perform one side of a bimanual task while a human performs the other, then swapping roles to cover both sides. RGB-D observations from a wrist-mounted and fixed camera are augmented into synthetic demonstrations for target bimanual robots using state-of-the-art hand pose estimation, image and point cloud segmentation, and inpainting. These synthetic demonstrations, grounded in real robot kinematics, are used to train bimanual policies. We evaluate MonoDuo on five tasks—box lifting, backpack packing, cloth folding, jacket zipping, and plate handover. Compared to approaches relying solely on human bimanual videos, MonoDuo enables zero-shot deployment on unseen bimanual robot configurations, achieving success rates up to 70\%. With only 25 target robot demonstrations, few-shot finetuning further boosts success rates by 65–70\% over training from scratch, demonstrating MonoDuo's effectiveness in efficiently transferring knowledge from single-arm robot data to bimanual robot policies. Project page: \url{https://bimanual-monoduo.github.io}

\end{abstract}

%%%%%%%%%%%%%%%%%%%%%%%%%%%%%%%%%%%%%%%%%%%%%%%%%%%%%%%%%%%%%%%%%%%%%%%%%%%%%%%%

\section{Introduction}
\label{sec:intro}
Bimanual robotic systems offer the potential to perform complex, coordinated manipulation tasks that are difficult or impossible for single-arm robots to execute. Many industrial and home tasks require two arms working in concert, with precise timing, spatial awareness, and physical coordination. However, a majority of available datasets and research infrastructure uses single-arm robots. This creates a bottleneck for learning bimanual policies, where the scarcity of bimanual robots significantly limits scalability.

We address this gap with \textit{MonoDuo}, a framework that democratizes bimanual robot learning by enabling training from single-arm demonstrations paired with human collaboration. MonoDuo builds on recent advances
in cross-embodiment learning--techniques for transferring
behaviors across different robot morphologies--and extends
them to the challenging setting of single-arm to bimanual
transfer. In our setup, a human teleoperates a single-arm robot to execute one side of a bimanual task while coordinating with a second human arm, alternating left/right roles across episodes to balance coverage. We then augment these trajectories into synthetic demonstrations for a target bimanual robot using hand pose estimation, image/point-cloud segmentation, and inpainting. To ensure both fidelity and coverage, MonoDuo introduces a novel, structured augmentation strategy that preserves real robot actions while filling in the missing arm with retargeted end-effector actions derived from the human collaborator. This structured mixing produces balanced datasets that maintain the action distribution when only a single-arm robot is available, while explicitly enabling cross-embodiment transfer for both human-to-robot and robot-to-robot generalization.

We evaluate MonoDuo on five challenging bimanual tasks—box lifting, backpack packing, cloth folding, jacket zipping, and plate handover. It achieves 35\%–70\% zero-shot success on unseen robot configurations. We additionally study a practical few-shot learning scenario with limited target demonstrations. In this setting, we show that MonoDuo improves sample efficiency significantly, increasing success rates by 65$\sim$70\% compared to policies without MonoDuo. This highlights our approach as both a transferable knowledge source and a complement to limited real-world bimanual data. This paper makes four contributions:

% We evaluate MonoDuo on five challenging bimanual tasks, including box lifting, backpack packing, cloth folding, jacket zipping, and plate handover. MonoDuo achieves zero-shot success rates of 35\%–70\% on unseen robot configurations. We additionally study a practical few-shot learning scenario, where only a small number of demonstrations on the target bimanual robot are available. In this setting, we show that MonoDuo improves sample efficiency significantly, increasing success rates by 65$\sim$70\% compared to policies without MonoDuo. This highlights the utility of our approach not only as a standalone source of transferable knowledge, but also as a complement to limited real-world bimanual data. This paper makes four contributions:

%This highlights how \textit{MonoDuo} can also complement limited real-world bimanual data. %This highlights the utility of our approach not only as a standalone source of transferable knowledge, but also as a complement to limited real-world bimanual data.

\begin{enumerate}[itemsep=2pt,leftmargin=*]
    \item \algoname, a novel framework for collecting demonstration data using one robot arm in collaboration with a human arm and synthesizing bimanual robot demonstrations.
    \item A data transformation pipeline that converts human-robot arm demonstrations into bimanual robot demonstrations using pose estimation, segmentation, and inpainting, and\textbf{ }introduces novel mixing of human retargeted end-effector actions with real robot actions to enhance policy learning in comparison to robot-only policies.
    \item Experiments suggesting that policies trained with \algoname can generalize zero-shot to previously unseen bimanual robot configurations, evaluated on a set of 5 bimanual tasks.
    \item Experiments suggesting that \algoname significantly improves sample efficiency when finetuned with 25 bimanual robot demonstrations in comparison to policies trained from scratch.  %, reducing the cost of real-world data collection;
    %\todo{add specific novel aspects or features} %We validate our approach across a set of challenging bimanual tasks, including lifting, handovers, zipping, unsealing containers, and coordinated actuation of mechanisms.
\end{enumerate}

%======================================

\section{RELATED WORKS}

\subsection{Learning-Based Approaches to Bimanual Manipulation}

Existing learning-based approaches to equip robots with bimanual manipulation skills can be broadly classified into three categories: learning from demonstrations~\cite{stepputtis2022system,grannen2023stabilize,zhao2023learning,fang2023low,cheng2024open,wu2024gello,iyer2024open,lin2024learning}, sim-to-real reinforcement learning~\cite{huang2023dynamic,lin2024twisting,lin2025sim}, and learning from human videos~\cite{xiong2021learning,bahl2022human,wang2023mimicplay,li2024okami,bahety2024screwmimic,zhu2024vision,zhou2025you} or human motion data~\cite{chen2024object,wang2024dexcap}. In learning from demonstrations, a human teleoperates the robot arms or end-effectors to directly collect sensorimotor data from the bimanual robot.
% In learning from demonstrations, a human controls the robot arms and end-effectors through kinesthetic teaching or a teleoperation system of choice, so that sensorimotor data can be directly collected from the bimanual robot. 
The collected data can then be used to train bimanual manipulation policies in a straightforward way, using state-of-the-art imitation learning policies~\cite{zare2024survey}. Advances in bimanual teleoperation systems~\cite{zhao2023learning,cheng2024open,wu2024gello,lin2024learning} and imitation learning algorithms~\cite{florence2022implicit,chi2023diffusionpolicy} in recent years have lowered the barrier for adopting this approach, making it a popular choice among both industry and academic labs. On the other hand, sim-to-real RL avoids real robot data by training policies on a digital twin of the target robot~\cite{huang2023dynamic,lin2024twisting,lin2025sim}, then transferring to hardware, but faces challenges in reward design and the sim-to-real gap.
% Sim-to-real reinforcement learning approaches, on the other hand, typically do not use any real-world human or robot data. Existing works~\cite{huang2023dynamic,lin2024twisting,lin2025sim} instead first learns bimanual policies using the ``digital twin'' of the target robot in a simulator of choice, then transfer the learned policy to the real bimanual robot. This often leads to challenges in reward design and sim-to-real gap.
Learning from human video or motion data leverages human action priors to ease bimanual policy learning but lacks directly usable robot actions. Our work is most related to this line, especially human video.
% Learning from human video or motion data falls between the previous two approaches, in that it eases bimanual policy learning by learning directly from human action priors but does not contain robot action data that can be directly used. Our work is most closely related to learning from human video, which we discuss below. %This is also an approach that is closely related to cross-embodiment robot learning---which we discuss below.

%======================================
\addtocounter{figure}{-1}
\begin{figure*}[t]
% \begin{center}
\includegraphics[width=\textwidth]{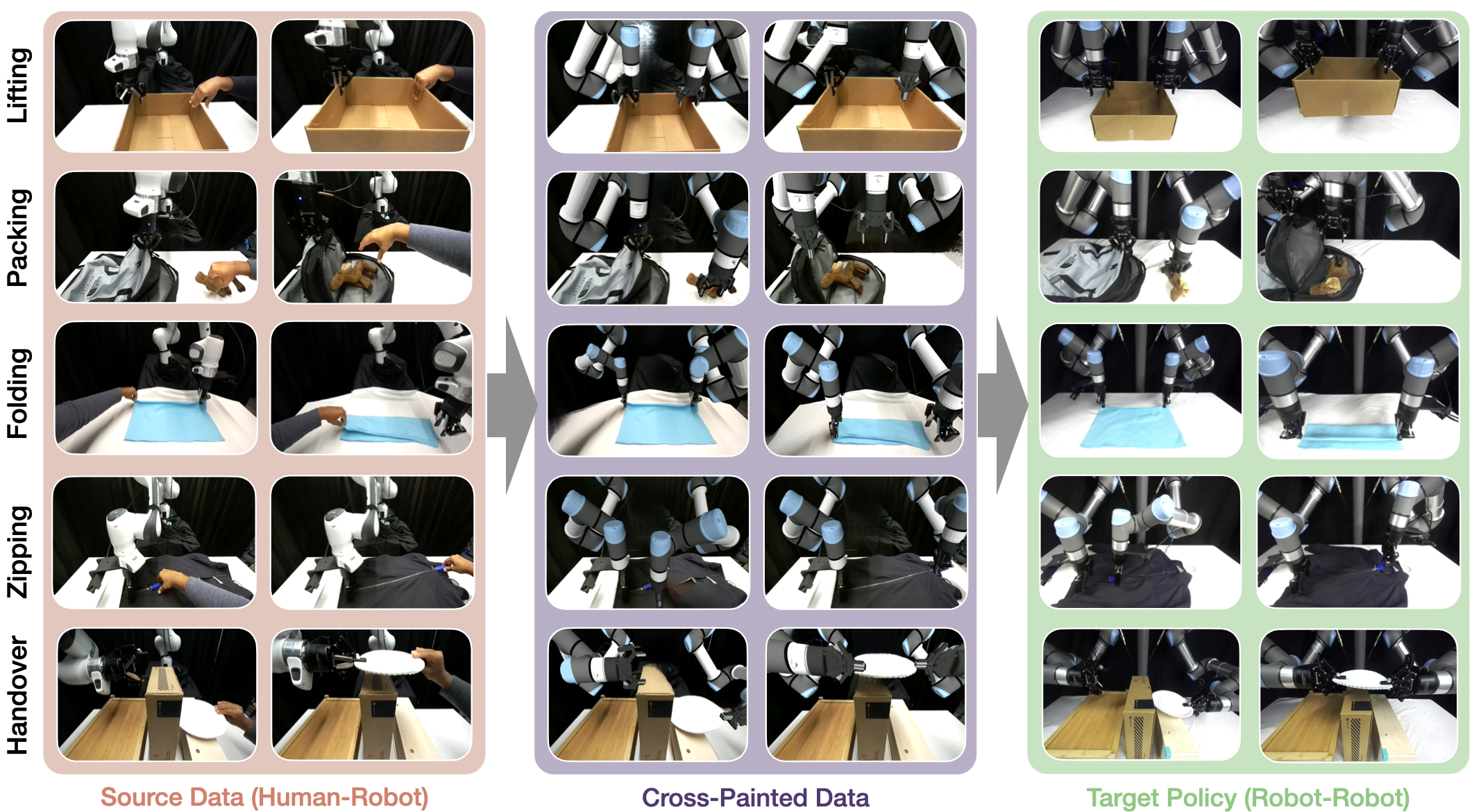}
% \end{center}
\caption{\textbf{From Human-Robot Demonstrations to Robot-Robot Policies.} Given collaborative demonstration trajectories between a single-arm robot and a human, MonoDuo uses state-of-the-art diffusion models to augment the image data and generate synthetic dataset tailored to a specified bimanual robot. Policies trained with the augmented dataset can be deployed on this target bimanual robot zero-shot. The same dataset can also be used to improve sample efficiency for few-shot learning.
}
\label{fig:hrrr}
\vspace{-1em}

\end{figure*}
%======================================

\subsection{Cross-Embodiment Robot Learning}

Cross-embodiment robot learning~\cite{open_x_embodiment_rt_x_2023} aims to learn or transfer policies across robots with different physical embodiments, enabling generalization across hardware without retraining for each configuration. Prior work tackles this via domain randomization and robot-conditioned policies~\cite{yu2023multi, chen2018hardware, shao2020unigrasp, xu2021adagrasp, wang2018nervenet, sanchez2018graph, pathak2019learning, huang2020one, kurin2020my} or by leveraging large-scale real robot datasets~\cite{depierre2018jacquard, kalashnikov2018qt, levine2018learning, acronym2020, shafiullah2023dobbe, fang2023rh20t, ebert2021bridge, walke2023bridgedata} to improve robustness and generalization~\cite{jang2022bc, brohan2023rt1, brohan2023rt2, jiang2022vima, shah2023gnm, shah2023vint, lynch2023interactive, shridhar2022cliport, stone2023moo, shridhar2022peract, reed2022a, radosavovic2022real, bharadhwaj2023roboagent, chen2023palix, driess2023palme}. RoVi-Aug~\cite{chen2024roviaug} introduces diffusion-based augmentation that replaces the robot in demonstration images and generates new camera views, producing synthetic demonstrations across robots and viewpoints. In contrast, Mirage~\cite{chen2024mirage} performs test-time image editing to create the illusion that a familiar robot is performing the task; this ``cross-painting'' decouples visual embodiment differences from control and enables zero-shot transfer between single-arm robots. We build on these cross-painting techniques. Shadow~\cite{lepert2024shadow} further simplifies cross-embodiment image editing by overlaying segmentation masks of the source and target robots on training and test images, aligning train-test distributions. Learning from human video is a related form of cross-embodiment robot learning: Phantom~\cite{lepert2025phantomtrainingrobotsrobots} trains robots without robot demos by converting human videos into robot-like observations, while EgoMimic~\cite{kareer2024egomimicscalingimitationlearning} co-trains on egocentric human videos and teleoperated robot demonstrations using cross-domain alignment. Masquerade~\cite{lepert2025masquerade} edits in-the-wild human videos to create synthetic bimanual robot demonstrations and co-trains policies for robust zero-shot transfer to unseen scenes. Other frameworks~\cite{jain2024vid2robot,jang2022bc,kedia2024oneshotimitationmismatchedexecution,wang2023mimicplay} also explore translating human videos into robot actions via learned correspondences.

\subsection{Learning Bimanual Manipulation with A Single-Arm Robot}

Bimanual manipulation poses challenges for cross-embodiment learning due to coordinated, high-dimensional actions. %Most visual augmentation techniques so far have been evaluated on single-arm tasks, and extending them to two-arm settings is non-trivial. 

Prior multi-embodiment studies avoid direct visual retargeting, instead using specialized methods: CrossFormer~\cite{Doshi24-crossformer} adds a bimanual action head, DexMimicGen~\cite{jiang2024dexmimicen} expands limited demos via simulation synthesis, and AnyBimanual~\cite{lu2024anybimanual} composes single-arm skills with task-level reasoning.
% Prior multi-embodiment studies usually avoid direct visual retargeting or handle bimanual robots with specialized architectures. For example, CrossFormer~\cite{Doshi24-crossformer} demonstrates a single policy controlling a bimanual robot across embodiments with a separate action head for bimanual robots. DexMimicGen~\cite{jiang2024dexmimicen} tackles bimanual learning by leveraging a small set of teleoperated demonstrations to seed large-scale synthetic trajectory generation in simulation, where the demonstrations need to adhere to dual-arm execution trajectories on the same robot. AnyBimanual~\cite{lu2024anybimanual} addresses bimanual learning by composing single-arm skills through a task-level reasoning and planning module, guided by a few task-specific bimanual demonstrations.

%Recent one-shot video imitation approaches have also begun addressing bimanual skills. ScrewMimic~\cite{bahety2024screwmimic} maps a single human demonstration video into a screw-based motion representation, then refines it through interactive exploration to bootstrap a robot’s two-handed skill from that single example.

In contrast, our approach learns end-to-end coordinated bimanual policies directly from synthetic demonstrations generated using only single-arm robot data paired with human interaction. The only closely related work that we have found is LfDT~\cite{kobayashi2023lfdt}, which uses human-robot interaction videos to learn dual-arm policies by learning a CycleGAN to transform human-robot images into robot-robot images. However, LfDT requires robot-robot target domain videos for training the CycleGAN~\cite{zhu2017unpaired} and is only validated on relatively simple tasks such as pushing. To the best of our knowledge, this work is the first to learn bimanual manipulation policies using only a single-arm robot, and to demonstrate success on complex, contact-rich tasks with zero-shot success, as shown in Section~\ref{sec:exp}.

%Despite recent advances, we find no prior work that applies visual retargeting or augmentation techniques specifically to bimanual demonstrations. In particular, none of the aforementioned approaches utilize robot-human interaction videos to generate training data for a dual-arm robot. Our work is thus novel in this regard: by capturing human–robot interaction demonstrations and converting them into synthetic two-arm robot experiences, we bridge human and robot embodiments for bimanual policy learning in a way that has not been shown before. This cross-embodiment visual augmentation for bimanual manipulation addresses a gap left by prior single-arm-focused augmentation and imitation techniques.

% \toru{No paper to date achieves true zero-shot bimanual control from only single-arm robot data. Closest attempts either (i) reuse a pretrained single-arm policy but still fine-tune on a handful of dual-arm trials, (ii) translate human-plus-robot demos into dual-arm data, or (iii) rely on dual-arm videos / tele-operation. }

%======================================

\section{Problem Statement}
\label{sec:problem}
As described in Figure~\ref{fig:overview}, \algoname collects a demonstration dataset $\mathcal{D}^{\mathcal{S}} = \{\tau_1^{\mathcal{S}}, \tau_2^{\mathcal{S}}, ..., \tau_n^{\mathcal{S}}\}$ consisting of $2N$ successful trajectories of a source robot-human pair $\mathcal{S} = (\mathcal{S}_r, \mathcal{S}_h)$ performing some task.
Each trajectory $\tau_i^{\mathcal{S}}  = (\{o_{1..H_i}^{\mathcal{S}}\}, \{p_{1..H_i}^{\mathcal{S}_r}\}, \{p_{1..H_i}^{\mathcal{S}_h}\}, \{a_{1..H_i}^{\mathcal{S}_r}\}, \{a_{1..H_i}^{\mathcal{S}_h}\})$, where $\{o_1^{\mathcal{S}}, ..., o_{H_i}^{\mathcal{S}}\}$ is a sequence of RGB-D camera observations, $\{p_1^{\mathcal{S}_r}, ..., p_{H_i}^{\mathcal{S}_r}\}$ is the sequence of corresponding robot state observations, $\{p_1^{\mathcal{S}_h}, ..., p_{H_i}^{\mathcal{S}_h}\}$ is the sequence of corresponding human hand state observations, $\{a_1^{\mathcal{S}_r}, ..., a_{H_i}^{\mathcal{S}_r}\}$ is the sequence of corresponding robot actions, and $\{a_1^{\mathcal{S}_h}, ..., a_{H_i}^{\mathcal{S}_h}\}$ is the sequence of corresponding retargeted end-effector actions derived from the human. 
Robot state observations include gripper pose and opening width, while human hand states come from a hand pose estimation algorithm. Each robot or human-retargeted action consists of gripper pose and opening width. To account for morphological differences between the human hand and a parallel-jaw gripper, \algoname includes a module that translates estimated human hand poses into gripper actions. Details on obtaining human-retargeted actions are provided in Section~\ref{sec:method}.
% The robot state observations consist of current gripper pose and opening width. The human hand state observations consist of parameters returned from a hand state estimation algorithm. Each robot action or human retargeted action consists of gripper pose and opening width. Since the human hand has a different morphology from a parallel-jaw gripper, \algoname includes a module to translate estimated human hand pose to gripper pose and opening width.
% We will elaborate more on how the human retargeted actions are obtained from estimated hand states in Section~\ref{sec:method}.

\algoname then augments $\mathcal{D}^{\mathcal{S}} $ into $\mathcal{D}^{\text{Aug}}$ to train a bimanual robot policy that can be deployed on a specified target bimanual robot $\mathcal{T}$ without test-time modification. This is illustrated in Figure~\ref{fig:hrrr}. We assume the grippers of robot $\mathcal{S}$ and robot $\mathcal{T}$ are both parallel-jaw grippers, and that each single-arm robot with gripper has kinematics that can be approximated with a human arm and hand. We also assume fixed and known camera poses for both the source and target domains. This allows us to render robots with known URDFs in ways that are within the training image distribution.
Similar to prior work~\cite{chen2024mirage, shah2023gnm, yang2023polybot, yang2024pushing}, we use Cartesian control and assume known inverse kinematics of the end-effector coordinate frames with respect to robot bases, such that we can use a rigid transformation $T_{\mathcal{T}}^{\mathcal{S}}$ to preprocess the data and align all end-effector poses $p^\mathcal{S} = T_{\mathcal{T}}^{\mathcal{S}} p^\mathcal{T}$ and actions $a^\mathcal{S} = T_{\mathcal{T}}^{\mathcal{S}} a^\mathcal{T}$ into the same vector space. Thus, for notational convenience, we omit the superscript differentiating end-effector poses and actions between  $\mathcal{S}$ and $\mathcal{T}$.

After data augmentation, we learn a policy $\pi(a_t | o_t^{\mathcal{T}}, p_t)$ on $\mathcal{D}^{\text{Aug}}$ using a behavior cloning algorithm of choice. At test time, this policy takes as inputs the observations from the target robot and outputs actions that can be deployed on the target robot. In a second set of experiments, we co-train $\mathcal{D}^{\text{Aug}}$ with a small number of demonstration data $\mathcal{D}^{\mathcal{T}}$ directly obtained from the target bimanual robot, and study how this leads to improvement on few-shot generalization.

%===============================================================================

\begin{figure*}[t]
\begin{center}
\includegraphics[width=\textwidth]{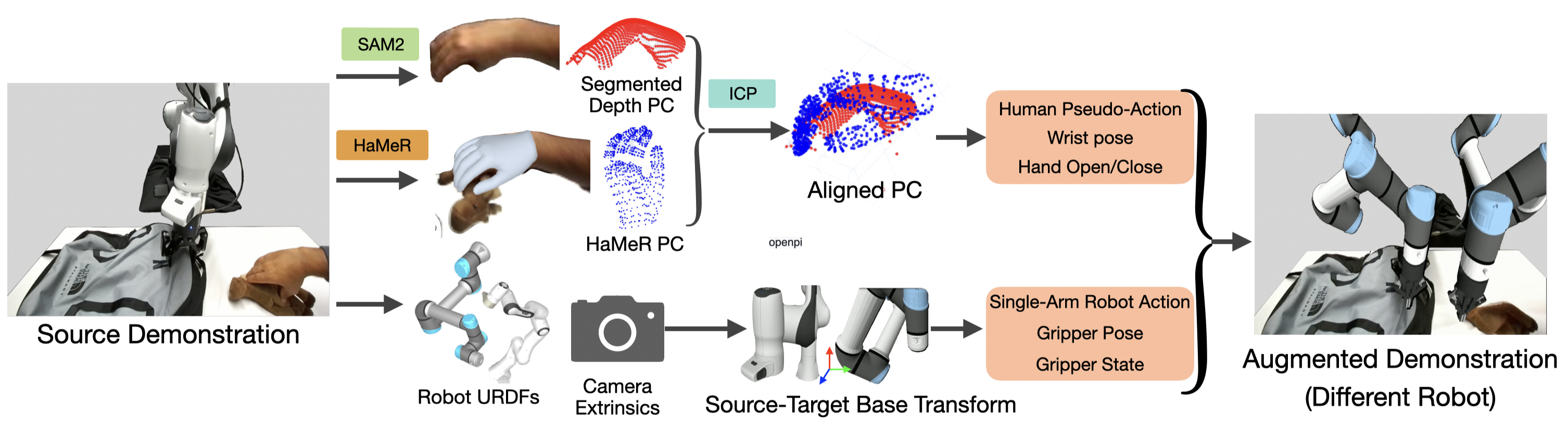}
\end{center}
\caption{\textbf{Data Collection and Dataset Augmentation.} \textit{Left}: We apply HaMeR~\cite{pavlakos2024reconstructing} to estimate the hand pose at each frame and refine with ICP~\cite{besl1992method,chen1992object}. The refined hand pose is then retargeted into robot end-effector actions in the source dataset. \textit{Right}: We perform cross-painting from both the source robot and the human arm to the target robot.}
\label{fig:pipeline}
\vspace{-1em}
\end{figure*}

\section{MonoDuo}

\label{sec:method}

In this section, we describe more details of how \algoname enables the learning of bimanual robot policies when only a single-arm robot is available. An overview is shown in Figure~\ref{fig:overview}.

\subsection{Data Collection}

For each bimanual task, we collect a source dataset $\mathcal{D}^{\mathcal{S}}$ using a human to teleoperate a single-arm robot to collaborate with a human partner on the task. To ensure a balanced data distribution, the roles of left-arm and right-arm are alternated across episodes. Specifically, human arm and robot collect $N$ trajectories on each side, where the total number of trajectories is $2N$. Data is collected in the format outlined in Section~\ref{sec:problem}.

We resolve the morphology gap between human and robot by retargeting the human arm-hand motions into end-effector actions. This is feasible based on two observations: (1) human wrist pose can be approximated as robot end-effector pose; (2) human hand pose can be approximated as gripper state. We begin by estimating the 3D human hand pose at each timestep---specifically, by applying HaMeR \cite{pavlakos2024reconstructing} to each RGB image from camera observation $o_t^{\mathcal{S}}$. HaMeR predicts 21 keypoints, $\hat{\mathbf{X}}_t \in \mathbb{R}^{21 \times 3}$, corresponding to anatomical landmarks following the MANO~\cite{romero2017mano} model. Since HaMeR struggles to estimate the absolute 3D pose due to its reliance on a monocular image, we incorporate depth to refine this estimate. In the RGB image observation, we use SAM2~\cite{ravi2024sam2} to obtain a segmentation mask of the hand; then, we extract a partial point cloud of the hand by applying the segmentation mask on the aligned depth image. Next, we align the HaMeR-predicted mesh $\hat{\mathbf{V}}_t$ with the segmented hand point cloud $\mathbf{P}_t$ via Iterative Closest Point (ICP) registration~\cite{besl1992method}, obtaining the optimal rigid transformation $\mathbf{T}_t \in SE(3)$ such that $\mathbf{P}_t \approx \mathbf{V}_t = \mathbf{T}_t \hat{\mathbf{V}}_t$. Since $\hat{\mathbf{V}}_t$ and $\hat{\mathbf{X}}_t$ are internally consistent, we can apply $\mathbf{T}_t$ to the predicted keypoints to refine their positions: $\mathbf{X}_t = \mathbf{T}_t \hat{\mathbf{X}}_t$.
%Since HaMeR models all hand joints as ball joints, it often predicts unrealistic finger configurations when hand keypoints are occluded in the RGB image. %—an issue exacerbated during grasping.
%To address this, we constrain the last two joints of the thumb and index fingers to a single degree of freedom, limiting their movement to anatomically feasible ranges. This ensures more accurate finger pose estimation when occlusions occur.
Once the keypoints $\mathbf{X}_t$ are refined, we define the human retargeted actions $a_{t}^{\mathcal{S}_h}$ in $\mathcal{D}^{\mathcal{S}}$ as follows: the end-effector pose is set as the estimated wrist pose, and the gripper opening is computed as a binary variable based on the scalar angle defined by three MANO~\cite{romero2017mano} landmarks: thumb fingertip, index finger fingertip, and index proximal frame. We set a threshold value for the scalar angle value, such that angle below which is translated to a closed gripper.% To mitigate slippage during grasping, we enforce a threshold, setting the bottom 20th percentile of predicted gripper distances in a single trajectory to fully closed.
%Since HaMeR predicts keypoints in the camera's reference frame, we convert the extracted hand pose and ... into the robot's frame using the known camera extrinsics of our target setup to obtain the final robot action $a_{r,t}$.

%======================================
\begin{figure*}[t]
\begin{center}
\includegraphics[width=\textwidth]{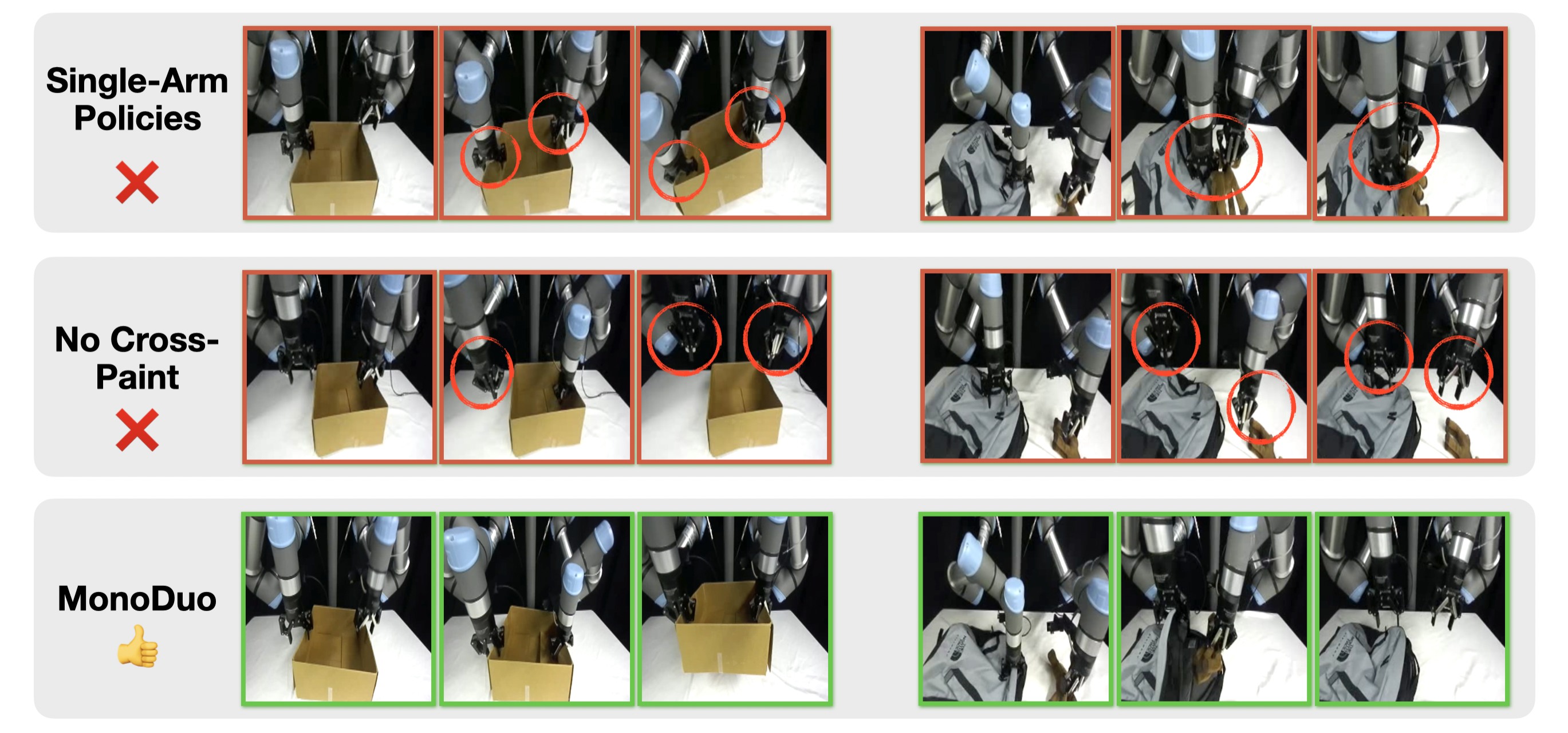}
\end{center}
\caption{\textbf{Examples of zero-shot rollout on the target bimanual UR5e.} \textit{Left:} Lift Box; \textit{Right:} Pack Bag. Single-Arm policies do not coordinate the actions well, leading to asynchronous movements as shown in the Lift Box task and collision in the Pack Bag task. Policies trained without cross-painting are less robust and misgrasps often. \algoname exhibits coordinated behaviors while being precise.}
\label{fig:capabilities}
\vspace{-1em}
\end{figure*}
%======================================

\subsection{Dataset Augmentation}
%These human-robot demonstrations are then augmented into synthetic robot-robot demonstrations using segmentation and inpainting techniques, creating a visually and physically grounded dataset for bimanual robots.

Given the source dataset $\mathcal{D}^{\mathcal{S}}$, we aim to augment it into $\mathcal{D}^{\text{Aug}}$ to learn a bimanual policy that can be deployed on the target bimanual robot $\mathcal{T}$. To this end, we apply ``cross-painting''---which in prior works~\cite{chen2024roviaug,chen2024mirage} means replacing the source robot with the target robot in the camera observations at test time so that it appears to the policy as if the source robot were performing the task. In this work, we extend cross-painting to also include human as a data source. We describe the details below, and illustrate the cross-painting procedure in Figure~\ref{fig:hrrr}.

% robot segmentation is analytic, assuming known camera calibration and using mujoco to render robot in images to figure out the mask (mirage)
% robot-to-robot and human-to-robot is analytic like mirage, we render target robot image using mujoco
% but data augmentation as Rovi-Aug, missing pixel inpainting is still E2FGVI
\paragraphc{Source Robot to Target Robot Cross-Painting.}
We leverage knowledge of the source and target robot URDFs and camera poses to perform robot-robot cross-painting at training time, as illustrated in Figure~\ref{fig:pipeline}. First, given known camera extrinsics, we re-project the images from the source domain to the target domain given that depth sensing is available. Next, given the RGB image observation and joint angles of the source robot, we use a renderer to determine which image pixels correspond to the source robot and mask out these pixels. Then, we inpaint the missing pixels using a video inpainting model E$^2$FGVI~\cite{liCvpr22vInpainting}. Finally, we use the URDF of the target robot to solve for the joint angles that would put its end effector at the same pose as that of the source robot, render it using a simulator, and overlay it onto the source image. For the gripper, we similarly compute and set the joints of the target robot gripper in the renderer so that its width would roughly match that of the source robot's gripper. To prevent the trained policy on the augmented data from overfitting to the synthetic robot visuals, we perform random brightness augmentation to the generated robot before pasting it. Previous work~\cite{chen2024roviaug} has demonstrated this random brightness augmentation to significantly help improve the performance of trained policies.

% hand segmentation is SAM2 without finetuning (phantom)
\paragraphc{Human to Target Robot Cross-Painting.} Cross-painting from human to target robot largely follows the same process as robot-robot cross-painting, except that we segment out the pixels corresponding to the human arm using SAM2~\cite{ravi2024sam2} before replacing the human embodiment with a robot. The target model is similarly rendered, with its end effector pose and gripper opening width corresponding to the human retargeted end-effector action extracted.

\subsection{Policy Training}

After applying dataset augmentation, we can train a policy $\pi$ based on the Diffusion Policy architecture~\cite{chi2023diffusionpolicy} on the augmented dataset ${\mathcal{D}^{Aug}}$ and zero-shot deploy the policy on the target robot $\mathcal{T}$. The policy input is RGB image observations and bimanual robot state observations; policy output is bimanual robot actions. For challenging tasks or when there is a large difference in the dynamics between the robots, we can also collect a small demonstration dataset $\mathcal{D}^{\mathcal{T}}$ on the target robot directly and few-shot finetune $\pi$ on $\mathcal{D}^{\mathcal{T}}$ to further improve policy performance.
% Alternatively, we can co-train $\pi$ on ${\mathcal{D}^{\mathcal{S}}}^{\text{Vi-Aug}} \bigcup {\mathcal{D}^{\mathcal{S}\rightarrow \mathcal{T}}}^{\text{Vi-Aug}}$ to obtain a multi-robot policy. Additionally, if we have multiple datasets with different tasks, can mix-and-match the datasets and train a multi-robot multi-task policy. For example, given data $\mathcal{D}_1^{\mathcal{S}}$ and $\mathcal{D}_2^{\mathcal{T}}$ with robot $\mathcal{S}$ performing task 1 and robot $\mathcal{T}$ performing task 2, we can train on the cross-product $\mathcal{D}_1^{\mathcal{S}} \bigcup {\mathcal{D}_2^{\mathcal{T}\rightarrow \mathcal{S}}} \bigcup {\mathcal{D}_1^{\mathcal{S}\rightarrow \mathcal{T}}} \bigcup \mathcal{D}_2^{\mathcal{T}}$ and their viewpoint-augmented versions to obtain a policy that can perform both tasks on both robots. In this way, we efficiently reuse the datasets and explicitly encourage transfer between robots and skills.

%===============================================================================

\section{Experiments}
\label{sec:exp}

\subsection{Hardware Setup and Task Definition}

We use a Franka arm as the single-arm source robot, and a pair of UR5e arms setup as the bimanual target robot. For RGB-D data collection, we use a ZED2 stationary fixed camera and a ZED-mini wrist-mounted camera.
We design five bimanual tasks for policy evaluation: (1) \textbf{Box Lifting}: the robot needs to coordinate the two grippers to lift up a box; (2) \textbf{Backpack Packing}: the robot needs to use one gripper to open a backpack, pick up a toy using the other gripper, put the toy into the backpack, and finally close the backpack with the first arm; (3) \textbf{Jacket Zipping}: the robot uses one arm to pin the jacket and the other arm to grasp and zip up the zipper of a jacket; (4) \textbf{Plate Handover}: the robot uses one gripper to pick up a plate and hands it over to the other gripper, while the other gripper needs to come to the waiting pose, stably grasp the plate, and put it down; (5) \textbf{Cloth Folding}: the robot needs to coordinate the two grippers to fold a piece of cloth by half. All tasks require highly coordinated behaviors of two arms and cannot be accomplished with a single-arm robot.

% \begin{enumerate}[itemsep=2pt,leftmargin=*]
%     \item box lifting: 
%     \item backpack packing: 
%     \item jacket zipping:
%     \item plate handover:
%     \item TBD:
% \end{enumerate}

\subsection{Implementation Details}

We collect 200 source-robot demonstrations per task, split evenly between left- and right-side human roles. Starting object positions and starting robot positions are randomized for each demonstration. For the generalization experiment, we additionally collect 25 trajectories on the target robot. On average, collecting a collaborative demonstration took 37 seconds, whereas a full teleoperated bimanual demonstration required $\sim$3 minutes, thus our collaborative setup reduced collection time by $\sim$79\%. Offline augmentation adds $\sim$106\,s per trajectory on an NVIDIA A100 GPU, and can be parallelized across demonstrations. For teleoperation, we use the Meta Quest device. We use the UNet-based Diffusion Policy as outlined in Chi et al.~\cite{chi2023diffusionpolicy}, with a ResNet encoder for visual observations. Policies take Cartesian proprioception, 2 image observations, and predict Cartesian end-effector actions.

\subsection{Results}

\paragraphc{Zero-Shot Bimanual Policies.} We report success rates of zero-shot deployed \algoname policies on each evaluation task in Table~\ref{tbl:baselines} and visualize their qualitative behaviors in Figure~\ref{fig:capabilities}. These results show that \algoname is able to effectively bridge both the visual and physical gaps among different robots and human, allowing one to learn bimanual policies when only a single-arm robot is available.

\paragraphc{Few-Shot \algoname.}  We study the finetuned performance of \algoname in comparison with bimanual policy training from scratch. We collect an additional 25 trajectories per task from direct teleoperation of the target bimanual UR5e robot. Using these 25 trajectories, we finetune \algoname and also train new policies from scratch on the same data. This situation corresponds to a common realistic scenario, where only a small number of demonstrations on the target bimanual robot is available. 
Results in Table~\ref{tab:fewshot} demonstrate that \algoname significantly improves learning efficiency in comparison to policy training from scratch, achieving higher performance with the same number of real demonstrations. Notably, few-shot \algoname boosts success rates on \textit{box lifting} from 30\% to 100\%, \textit{backpack packing} from 25\% to 90\%, and \textit{jacket zipping} from 5\% to 75\%, highlighting our approach's ability to complement limited real-world bimanual data.

\begin{table*}[t]
\vspace*{10pt}
\centering
\small
\setlength{\tabcolsep}{4pt}
\begin{tabular}{c|>{\footnotesize}c>{\footnotesize}c>{\footnotesize}c>{\footnotesize}c|>{\footnotesize}c>{\footnotesize}c>{\footnotesize}c>{\footnotesize}c>{\footnotesize}c}
\toprule
\multirow{3}{*}{\textbf{Policies}} 
& \multicolumn{4}{c|}{\textbf{Policy Attributes}} 
& \multicolumn{5}{c}{\textbf{Task Success Rates}} \\
& \makecell{Use\\Robot} & \makecell{Cross\\Paint} & \makecell{Retargeted\\Action} & \makecell{Weight\\Sharing}
& \makecell{Lift\\Box} & \makecell{Pack\\Bag} & \makecell{Zip\\Jacket} & \makecell{Handover\\Plate} & \makecell{Fold\\Cloth} \\
\midrule

Robot-Only Policies (Naive)     & \cmark & \cmark &      &        & 15\% & 10\%  & 15\%  & 0\%  & 5\% \\
Pure Human Videos               &       & \cmark & \cmark & \cmark & 10\% & 0\%  & 0\%  & 0\%  & 0\% \\

Ablation: No Cross-Paint        & \cmark &       & \cmark & \cmark & 40\% & 30\% & 15\% & 10\%   & 15\% \\
Ablation: No Retargeted Action      & \cmark & \cmark &       & \cmark & 50\% & 20\% & 20\%  & 30\%   & 25\% \\
\algoname                       & \cmark & \cmark & \cmark & \cmark & \textbf{70\%} & \textbf{55\%} & \textbf{45\%} & \textbf{35\%} & \textbf{35\%} \\

\bottomrule
\end{tabular}
\vspace{5pt}
\caption{\textbf{Zero-shot experiments comparing \algoname with baselines.} Each policy is evaluated on five manipulation tasks in a zero-shot transfer setting from Franka-human demos to a bimanual UR5e. We collect 200 Franka-human demos per task. Most of MonoDuo's failures are due to missed grasps. In contrast, the baseline policies have a large number of two-arm coordination failures on all tasks.
}
\label{tbl:baselines}
\vspace*{-10pt}
\end{table*}

%======================================
% \begin{table}[t]
%     \centering
%     \small
%     \begin{tabular}{lcc}
%         \toprule
%         &  \textit{Scratch} & \textit{Few-Shot MonoDuo} \\ 
%         \midrule
%         Lift Box & 30\% & 100\% \\
%         Pack Bag & 25\% & 90\% \\
%         Zip Jacket & 5\% & 75\% \\
%         \bottomrule
%     \end{tabular}
%     \vspace{5pt}
%     \caption{\textbf{Comparing Few-Shot MonoDuo with Learning from Scratch.} MonoDuo helps improve sample efficiency significantly, as shown by the superior success rates compared to baseline after training both with the same number of 25 demonstrations.}
%     \label{tab:fewshot}
% \end{table}
%======================================

\paragraphc{Failure Mode Analysis.}
Across 100 zero-shot rollouts of MonoDuo, most failures come from grasping errors (71.2\%), including missed grasps, drops, and over-grasps. This is the primary bottleneck in \textit{Pack Bag}, \textit{Zip Jacket}, and \textit{Fold Cloth}. The failure distribution suggests MonoDuo is primarily limited by contact-sensitive execution, rather than task-level sequencing. Coordination issues are less common overall (9.6\%) but concentrate in tasks requiring tight timing and obstacle-aware motion (e.g., \textit{Lift Box} coordination, \textit{Plate Handover} collisions). The remaining 19.2\% of errors are task-specific mechanics, including \textit{Zip Jacket} pin failures, \textit{Fold Cloth} getting stuck, and \textit{Plate Handover} final placement errors, making \textit{Plate Handover} the most multi-stage-limited task (46.2\% non-grasp failures).

\paragraphc{Comparison with Baselines.} We evaluate \algoname against 4 baselines: (1) \textit{Robot-Only Policies}: Two independent single-arm policies, each conditioned on the cross-painted global observation and its respective robot state, trained to predict the action for a single arm. (2) \textit{Pure Human Videos}: A policy trained solely on bimanual human-only demonstrations, using cross-painting to simulate robot embodiment and pose estimation of both hands. (3) \textit{No Cross-Paint}: An ablation that removes the visual domain alignment step, training instead on raw images while still leveraging both human and robot action supervision. (4) \textit{No Retargeted Actions:} Similar to \algoname in using a unified policy architecture, but excludes human retargeted actions during training, relying only on robot action supervision. Quantitative results in Table~\ref{tbl:baselines} show that all three core components of \algoname—joint human-robot demonstrations, robot-robot cross-painting, and human-robot cross-painting—are essential for learning effective bimanual coordination policies. Figure~\ref{fig:capabilities} show some qualitative examples, and we analyze key insights from ablation studies below.

\begin{table}[t]
    \centering
    \small
    \setlength{\tabcolsep}{4pt}
    \begin{tabular}{lcc}
        \toprule
                 & \textit{No WristCam} & \textit{With WristCam} \\ 
        \midrule
        Lift Box     & 60\%  & \textbf{70\%} \\
        Pack Bag     & 40\%  & \textbf{55\%} \\
        Zip Jacket   & 25\%  & \textbf{45\%} \\
        \bottomrule
    \end{tabular}
    \caption{\textbf{Impact of Wrist Camera on Zero-Shot Performance.} Using only a third-person camera yields strong results, but wrist-mounted cameras improve precision in tasks requiring fine manipulation, such as zipper grasping.}
    \label{tab:wristcam}
    \vspace{-10pt}
\end{table}

\begin{table}[t]
    \centering
    \small
    \setlength{\tabcolsep}{4pt}
    \begin{tabular}{lcc}
        \toprule
                 & \textit{Scratch} & \textit{Few-Shot MonoDuo} \\ 
        \midrule
        Lift Box    & 30\%  & \textbf{100\%} \\
        Pack Bag    & 25\%  & \textbf{90\%} \\
        Zip Jacket  & 5\%   & \textbf{75\%} \\
        \bottomrule
    \end{tabular}
    \caption{\textbf{Few-Shot Learning with MonoDuo.} Incorporating 25 bimanual robot demonstrations enables MonoDuo to significantly outperform policies trained from scratch, demonstrating improved sample efficiency.}
    \label{tab:fewshot}
    \vspace{-10pt}
\end{table}

\paragraphc{Importance of Weight-Sharing.} Our results indicate that using two disjoint robot-only policies, even when paired with cross-painted visual input, fails to produce reliable coordination. Each arm tends to execute its part of the task independently, leading to asynchronous behavior. While some tasks may succeed occasionally, the overall quality is poor—for example, lifting a box unevenly—and no success is observed in tasks demanding precise temporal synchronization. 

\paragraphc{Importance of Cross-Painting.} Removing cross-painting results in a 20–30\% drop in success across all tasks, demonstrating the critical role of visual domain alignment in enabling effective transfer. Cross-painting helps mitigate the embodiment gap and enables the model to generalize better. 

% \paragraphc{Value of Human Retargeted Actions.} Training solely on human video data yields near-zero success in zero-shot settings, primarily due to noisy hand pose estimation. To enable a fair comparison, we also include results of  \algoname trained without wrist cameras in Table~\ref{tab:wristcam}. \algoname addresses this limitation through structured augmentation: it preserves real robot actions for high-fidelity supervision while incorporating human retargeted actions to balance the training distribution. This reduces reliance on perfect human pose estimation, enhances policy robustness, and enables fair cross-embodiment transfer. Empirically, combining human  retargeted actions with robot actions improves performance across tasks, even when hand pose estimation is imperfect, demonstrating that leveraging both sources produces more reliable bimanual policies than using human data alone.

\paragraphc{Value of Human Retargeted Actions.} Training solely on human video yields near-zero zero-shot success primarily due to noisy hand pose estimation. For a fair comparison, we also report \algoname without wrist cameras in Table~\ref{tab:wristcam}. \algoname mitigates this via structured augmentation that preserves real robot actions for high-fidelity supervision while adding human retargeted actions to balance the training distribution. This reduces reliance on perfect pose estimation and enables fair cross-embodiment transfer. Empirically, combining human retargeted and robot actions improves performance across tasks, yielding more reliable bimanual policies than human data alone.

%
%
%
% \paragraphc{Value of Human Pseudo-Actions.\} Training solely on human video data yields near-zero success in zero-shot settings, primarily due to noisy hand pose estimations. To enable a fair comparison, we also include results of \algoname trained without wrist cameras in Table\~\ref\{tab:wristcam\}. \algoname addresses this limitation through structured augmentation: it preserves real robot actions for high-fidelity supervision while incorporating pseudo-actions from the human arm to balance the training distribution. This reduces reliance on perfect human pose estimation, enhances policy robustness, and enables fair cross-embodiment transfer. Empirically, combining human pseudo-actions with robot actions improves performance across tasks, even when hand pose estimation is imperfect, demonstrating that leveraging both sources produces more reliable bimanual policies than using human data alone.

\section{Conclusion}
\label{sec:conclusion}

% We present \algoname, a novel framework for learning bimanual robot policies using only demonstrations from a single-arm robot in collaboration with a human. By alternating roles between human and robot across episodes and applying vision-based augmentation techniques, \algoname generates synthetic bimanual demonstrations tailored to a specified target robot. This approach enables training policies that generalizes zero-shot to previously unseen bimanual robot configurations, and significantly improves sample efficiency in low-data regimes. We validate \algoname on five challenging bimanual manipulation tasks, demonstrating its effectiveness and superior performance over baselines. We believe \algoname can be a scalable and accessible solution for bimanual robot learning.
We present \algoname, which learns bimanual policies from single-arm human-robot demonstrations by alternating roles and applying vision-based augmentation to synthesize target-robot bimanual data. \algoname generalizes zero-shot to unseen configurations and improves sample efficiency in low-data regimes. We validate \algoname on five challenging bimanual tasks, showing superior performance over baselines, highlighting its potential as a scalable solution for bimanual robot learning.

\section{Limitations and Future Work}
\label{sec:limitations}
While \algoname provides a scalable framework for learning bimanual policies from a single-arm robot, several limitations remain. First, the approach assumes fixed and known camera calibration across domains, simplifying rendering and cross-painting but limiting use in uncalibrated settings. Second, it requires depth sensing for 3D hand pose refinement and segmentation, necessitating an RGB-D camera. MonoDuo’s mix of human-retargeted and real robot actions mitigates some challenges, but the augmentation pipeline remains sensitive to noise in hand-pose estimation and inpainting, especially under occlusion or poor lighting. However, our modular design allows integrating newer, improved hand-pose, segmentation, and inpainting models to enhance retargeted action quality and overall policy performance, potentially improving robustness in future work.

\renewcommand*{\bibfont}{\footnotesize}% \setlength{\bibitemsep}{0pt}
\printbibliography
% \bibliographystyle{plainnat}
% \bibliography{references}        % main.bib file (no .bib extension)
% \printbibliography

\end{document}